\documentclass[conference]{IEEEtran}
\IEEEoverridecommandlockouts
\usepackage{cite}
\usepackage{amsmath,amssymb,amsfonts}
\usepackage{algorithmic}
\usepackage{graphicx}
\usepackage{textcomp}
\usepackage{xcolor}

\usepackage{hyperref}
\hypersetup{hypertex=false,
            colorlinks=true,
            linkcolor=black,
            anchorcolor=black,
            citecolor=black}
\usepackage{cite}
\renewcommand{\thesection}{\textbf{\Roman{section}}}
\renewcommand{\thetable}{\arabic{table}} % 确保表格编号为阿拉伯数字

\usepackage[numbers,sort&compress]{natbib}

\usepackage{amsmath,amssymb,amsfonts}
\usepackage{algorithmic}
\usepackage{graphicx}
\usepackage{textcomp}
\usepackage{xcolor}
\usepackage{tabu}                     % 表格插入
\usepackage{multirow}                 % 一般用以设计表格，将所行合并
\usepackage{multicol}                 % 合并多列
\usepackage{multirow}                % 合并多行
\usepackage{amssymb}   % For checkmark symbol
\usepackage{array}     % For advanced table features
\usepackage{float}                    % 图片浮动
\usepackage{makecell}                 % 三线表-竖线
\usepackage{booktabs}                 % 三线表-短细横线
\usepackage{colortbl} % 在文档前言部分添加
\usepackage{etoolbox}

\usepackage{arydshln} % 添加虚线

\usepackage{tabularx}
\usepackage{booktabs} % For formal tables
\usepackage{siunitx} % Provides the S column type

\newcommand{\rz}[1]{\textcolor{black}{#1}}

\newcolumntype{C}[1]{>{\centering\arraybackslash}m{#1}}
\newcolumntype{R}{>{\raggedleft\arraybackslash}S} % 新的列类型定义，使红色文本居中
\usepackage{caption}

\usepackage{threeparttable} % 导入宏包 表格中加注脚

\renewcommand{\tablename}{Table} % To make 'Table' label uppercase
\def\BibTeX{{\rm B\kern-.05em{\sc i\kern-.025em b}\kern-.08em
    T\kern-.1667em\lower.7ex\hbox{E}\kern-.125emX}}
\begin{document}

\title{InfoSyncNet: Information Synchronization Temporal Convolutional Network for Visual Speech Recognition
% {\footnotesize \textsuperscript{*}Note: Sub-titles are not captured for https://ieeexplore.ieee.org  and
}
% \author{\IEEEauthorblockN{Anonymous Authors}} 
% \author{
% Yifang Xu\textsuperscript{1}, 
% Yunzhuo Sun\textsuperscript{2}, 
% Benxiang Zhai\textsuperscript{1}, 
% Youyao Jia\textsuperscript{3}, 
% and Sidan Du\textsuperscript{1} \\
% \textsuperscript{1}\textit{School of Electronic Science and Engineering, Nanjing University, China} \\
% \textsuperscript{2}\textit{School of Physics and Electronics, Hubei Normal University, China} \\
% \textsuperscript{3}\textit{Gosuncn Chuanglian Technology Co., Ltd., China} \\
% Email: \{xyf, zbx\}@smail.nju.edu.cn; sunyunzhuo98@outlook.com; coff128@nju.edu.cn
% }

\author{
\IEEEauthorblockN{
Junxiao Xue\textsuperscript{1},
Xiaozhen Liu\textsuperscript{1*}\thanks{*Corresponding author: liuxiaozhen123@gs.zzu.edu.cn},
Xuecheng Wu\textsuperscript{2},
Fei Yu\textsuperscript{3},
Jun Wang\textsuperscript{3}
}
\IEEEauthorblockA{
\textsuperscript{1}\textit{School of Cyber Science and Engineering,
Zhengzhou University, Zhengzhou, Henan, China
}}
\IEEEauthorblockA{
\textsuperscript{2}\textit{School of Computer Science
and Technology,
Xi’an Jiaotong University, Xi’an, Shaanxi, China
}}
\IEEEauthorblockA{
\textsuperscript{3}\textit{Research Center for Space
Computing System,
Zhejiang Lab, Hangzhou, Zhejiang, China
}}
\IEEEauthorblockA{
Emails: xuejx@zzu.edu.cn, liuxiaozhen123@gs.zzu.edu.cn,\\
wuxc3@stu.xjtu.edu.cn, yufei\_hitcs@163.com, wangjun@zhejianglab.org
}
}

% \vspace{-6mm}  % 适当压缩标题与正文之间的空白

% \and
% \and
% \and
% \and
% \and

\maketitle

\begin{abstract}
Estimating spoken content from silent videos is crucial for applications in Assistive Technology (AT) and Augmented Reality (AR). However, accurately mapping lip movement sequences in videos to words poses significant challenges due to variability across sequences and the uneven distribution of information within each sequence. To tackle this, we introduce InfoSyncNet, a non-uniform sequence modeling network enhanced by tailored data augmentation techniques. Central to InfoSyncNet is a non-uniform quantization module positioned between the encoder and decoder, enabling dynamic adjustment to the network's focus and effectively handling the natural inconsistencies in visual speech data. Additionally, multiple training strategies are incorporated to enhance the model's capability to handle variations in lighting and the speaker’s orientation. Comprehensive experiments on the LRW and LRW1000 datasets confirm the superiority of InfoSyncNet, achieving new state-of-the-art accuracies of 92.0\% and 60.7\% Top-1 ACC
\footnote{\href{https://github.com/liuxiaozhen123/InfoSyncNet}{https://github.com/liuxiaozhen123/InfoSyncNet}
}.
\end{abstract}


\section{INTRODUCTION}

Visual Speech Recognition (VSR), or lip reading, is the challenging process of interpreting spoken content solely from lip movements in silent video footage. This technology boasts widespread applications across multiple sectors, including information security, generating subtitles for silent films, facilitating communication for individuals who are deaf or mute \cite{ke2018}, driver-assistive systems \cite{ryumin2024audio}, and deducing spoken content in acoustically challenging environments \cite{burchi2023audio}.
\begin{figure}[thb]
    \centering
    \includegraphics[width=1.0\linewidth]{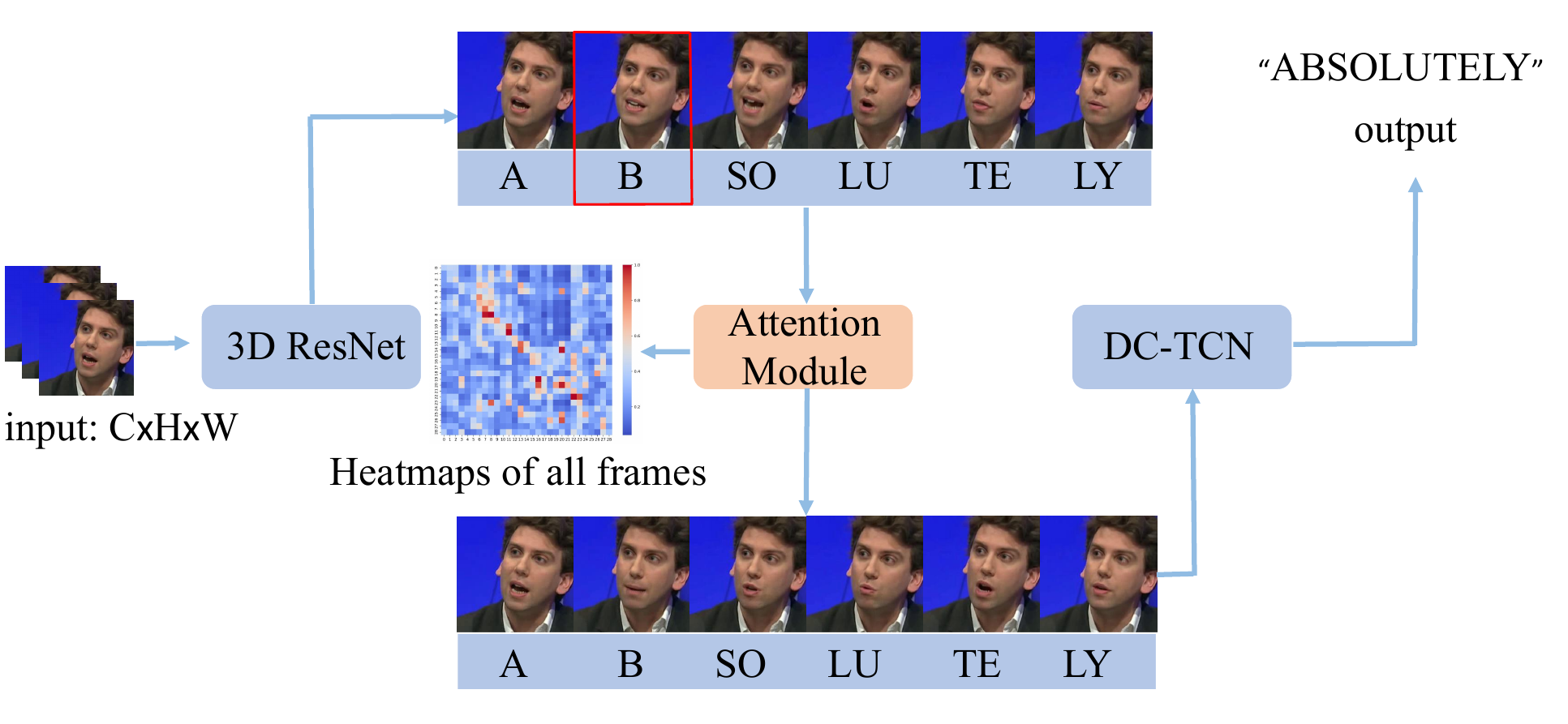}
    \caption{The basic idea is to insert a six-layer Transformer-based attention module, referred to as non-uniform quantization, between the front-end visual encoder and the back-end sequence decoder. The model selects frames evenly from the divided sequence for the first row, where the phoneme `B' in the red frame does not match the visual information. For the third row, key frames are selected based on the heatmap generated by the non-uniform quantization module, with higher heat values indicating key frames. This allows the model to focus on finer details in the lip reading sequence, such as the phoneme `B', which better aligns with actual lip movements.}
    \label{fig0}
    \vspace{-3mm} % 减少表格内容和下文之间的间距
\end{figure}

Significant advances in lip reading have been achieved with the introduction of large-scale datasets for training \cite{russakovsky2015imagenet} and the rapid evolution of deep learning models \cite{szegedy2015going}. However, learning the mapping between lip movements and spoken content is challenging for two main reasons. First, the uneven distribution of information in input sequences makes networks designed under the assumption of uniform information distribution struggle with this issue. Unlike classical sequence modeling tasks where the information load of sequence elements can be assumed consistent, lip movement sequences can vary significantly due to different speaking habits and camera exposure constraints. Second, variations in lighting and speaker orientation, resulting from capturing lip movements in diverse conditions, pose further challenges.

To effectively address the aforementioned challenges, various strategies have been implemented. Notably, Aldeneh Z et al. \cite{aldeneh2023role} conducted experiments to evaluate the impact of exaggerated and insufficient lip movements on visual speech quality, finding a pronounced preference for exaggerated movements which directly tackles the issue of uneven information distribution in lip movements due to different speaking habits. Kim et al.  \cite{kim2022speaker} developed a user-dependent padding framework tailored for lip shape recognition, which adapts to the variability in speaker characteristics, further addressing the non-uniform information distribution. A common technique in lip reading research \cite{feng2020learn} involves converting color images to grayscale, a practice that mitigates the influence of varying lighting conditions on recognition accuracy. Moreover, Feng et al.  \cite{feng2020learn} advanced the capabilities of ResNet-18 by incorporating the Squeeze-and-Excitation (SE) block \cite{hu2018squeeze}, markedly improving the convolutional layers' ability to discern and analyze key features from lip reading sequences, thus addressing challenges related to camera exposure and lighting variations.

Furthermore, adopting robust sequence modeling techniques is crucial for overcoming these challenges. Afouras T et al. \cite{afouras2018deep} introduced Transformers into visual speech recognition tasks, establishing a framework for subsequent research. The adoption of Temporal Convolutional Networks (TCNs) for backend sequence modeling has proven highly effective in recent approaches. Models like the Multi-Scale Temporal Convolutional Network (MS-TCN) and the Densely Connected Temporal Convolutional Network (DC-TCN) have been successfully applied in this field \cite{bai2018empirical,martinez2020lipreading,ma2022training,ma2021lip}, offering faster convergence and enhanced performance due to their fully convolutional architectures. The DC-TCN \cite{ma2022training} is regarded as the state-of-the-art model for word-level lip reading recognition. However, the dense connections \cite{huang2017densely} in the DC-TCN impair its ability to handle non-uniform information because they may reduce the network’s flexibility in dynamically adapting to varied data features.

In this paper, we propose a non-uniform sequence modeling network with specifically designed data augmentation techniques tailored for the VSR task to address the challenges outlined. Specifically, we introduce a non-uniform quantization module between the encoder network, which perceives local motion information, and the decoder network, which is sensitive to temporal dynamics, to model non-uniform sequences. As shown in Fig. \ref{fig0}, the first row evenly selects frames from the divided sequence, while the third row chooses key frames based on the heatmap from the non-uniform quantization module. This approach allows the model to focus on finer details of lip movements, such as the phoneme `B', which better aligns with practical standards, thereby effectively handling both cross-sequence diversity and intra-sequence non-uniformity. Lip movements can be considered as signals where the probability distribution of signal amplitudes is non-uniform, with small signals occurring much more frequently than large signals. Models designed under the assumption of uniformity often lead to information asynchrony. This issue can be alleviated by introducing an attention mechanism \cite{vaswani2017attention} at a highly abstract feature level, which selectively focuses on the most relevant parts of the sequence to enhance the interpretation of diverse lip movement data and improve information synchronization in non-uniform lip reading sequences. This targeted approach allows our proposed network to dynamically adjust its focus, enhancing the model’s ability to handle the natural inconsistencies found in visual speech data. 

Our work makes three main contributions:
% , the implementation code will be released alongside the paper upon its acceptance :
\begin{itemize}
\item We introduce a new model, known as \textbf{InfoSyncNet}, which enhances the handling of non-uniform information in lip movement sequences by applying an attention module to the spatial features. After extracting features through the visual front-end, spatial features for each frame are obtained via global average pooling. By calculating the correlation between frames, the approach addresses both inter-sequence diversity and intra-sequence unevenness.
\item We compare several VSR-specific training strategies on InforSyncNet. They include techniques such as Time Masking, Mixup, Word Boundary, and Label Smoothing.
\item Our results establish new state-of-the-art performance on major lip reading datasets, achieving accuracies of 92.0\% on the LRW dataset and 60.7\% on the LRW1000 dataset.
\end{itemize}
% First, the \textbf{InfoSyncNet} enhances the handling of non-uniform information in lip movement sequences through mutual information operations on the spatial features. Second, we compare several VSR-specific training strategies on InforSyncNet. Third, our results establish new state-of-the-art performance on major lip reading datasets, achieving accuracies of 92.0\% on the LRW dataset and 60.7\% on the LRW1000 dataset.

\begin{figure*}[thb]
    \centering
    \includegraphics[width=0.95\linewidth]{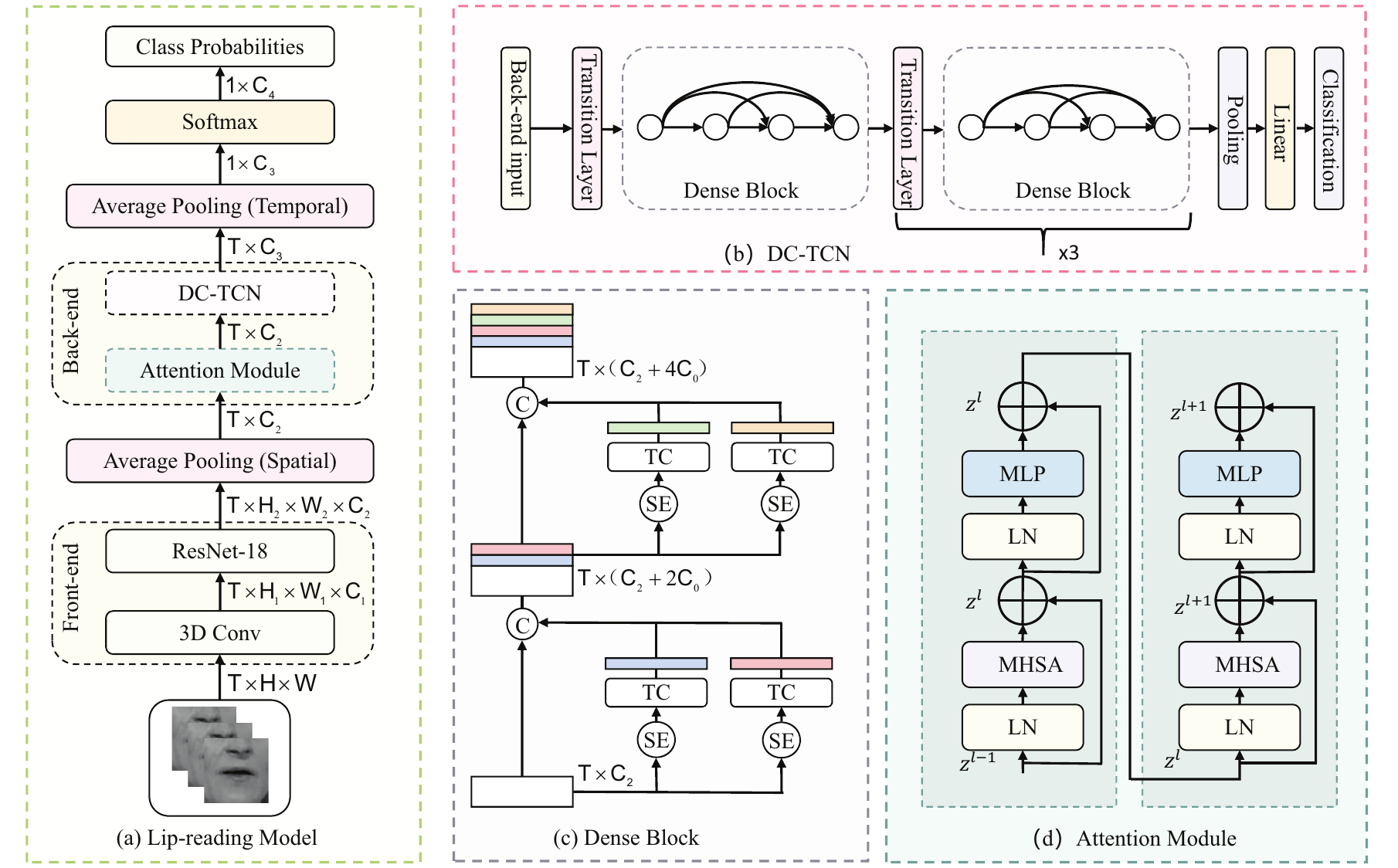}
    \caption{\rz{(a): The overall illustrations of our InfoSyncNet. We propose the attention-enhanced Dense Convolutional Network (DC-TCN) as the temporal model based on features extracted by modified ResNet-18 \cite{hara2017learning}.
    (b): DC-TCN. It consists of four dense blocks, each with three dense layers. (c): Dense block. SE: Squeeze-and-Excitation block \cite{hu2018squeeze}. TC: Temporal Convolutional block. (d): Attention Module. MHSA represents the Multi-Head Self-Attention mechanism.}}
    \label{fig1}
    \vspace{-3mm} % 减少表格内容和下文之间的间距
\end{figure*}

\section{BACKGROUND AND RELATED WORK}
Our work focuses on word-level lip reading, which aims to map sequences of lip movement images to corresponding words. Recently, this has been typically achieved using neural network architectures that consist of a front-end visual encoder and a back-end temporal decoder. The visual encoder captures local motion patterns, including frame-level and clip-level features, while the temporal decoder analyses patterns across the entire sequence.

% Traditional sequence modeling like Gaussian process dynamical models \cite{deena2010visual}, decision tree models \cite{kim2015decision}, and hidden Markov models \cite{anderson2013expressive} were widely used in solving visual speech recognition tasks. 

\textbf{VSR Visual Encoder.} In the initial stage, a feature extractor like Discrete Cosine Transform (DCT)~\cite{hong2006pca} was applied to the mouth region of interest (ROI). Traditional lip-reading models, such as Hidden Markov Models \cite{anderson2013expressive}, excel in temporal sequence modeling with simplicity but are limited by the state independence assumption and insufficient capacity to capture complex visual patterns. Chung et al. \cite{chung2017lip} proposed the first end-to-end deep visual representation learning for word-level VSR, using the VGG-M network to recognize words spoken by individuals from video sequences. Since its introduction in 2016, the 3DCNN \cite{ji20123d} has functioned as a spatial-temporal feature extractor, efficiently capturing valuable information across both spatial and temporal dimensions simultaneously. The 3D Convolutional layer and 2D ResNet \cite{hara2017learning} became the preferred choice for front-end visual encoding tasks, supported by significant contributions in the literature~\cite{axyonov2024audio,petridis2018end,weng2019learning}. Cheng et al.~\cite{cheng2020towards} proposed a method to enhance the performance of visual speech recognition in extreme poses. Ma et al.~\cite{ma2021towards} advocated for the deployment of the ShuffleNet v2 network~\cite{ma2018shufflenet} to streamline the front end. Subsequently, Koumparoulis et al.~\cite{koumparoulis2022accurate} adopted the EfficientNetV2 network~\cite{tan2021efficientnetv2} to further reduce the front-end’s complexity. Although these methodologies~\cite{ma2021towards, koumparoulis2022accurate} decreased computational demands, they also marginally impaired the accuracy of recognition.

\textbf{VSR Temporal Decoder.} For backend models, RNN-based models, like Long Short-Term Memory (LSTM) networks, have been widely employed to capture both global and local temporal dynamics \cite{xu2020discriminative}. Additionally, Bi-directional Gated Recurrent Units (BGRUs) have also achieved notable success \cite{feng2020learn}. In related fields, Lin WC et al. \cite{lin2023sequential} employed a bidirectional LSTM network to more comprehensively capture local emotional variations within sentences, effectively managing the uneven distribution of sequence information in speech emotion. Compared to RNN-based architectures, Transformers \cite{vaswani2017attention} have significant advantages in long-term dependency and parallel computation. Inspired by the tremendous success of transformer architectures in the NLP domain, researchers have recently applied transformers to computer vision tasks. Afouras T et al. \cite{afouras2018deep} were the first to introduce transformers into visual speech recognition tasks. Since the basic transformer does not specifically address short-term dependency, Zhang et al. \cite{zhang2019spatio} proposed multiple Temporal Focal blocks (TF-blocks) to help features focus on their neighbors and capture more short-term temporal dependencies. In recent years, TCN \cite{lea2017temporal} with dilated and causal convolutions, along with their variants such as MS-TCN and DC-TCN, have been effectively applied in this field \cite{bai2018empirical,martinez2020lipreading,ma2022training,ma2021lip}. These models, leveraging fully convolutional architectures, demonstrate faster convergence and superior performance compared to RNNs and Transformer-based models. 

\textbf{Training Strategies.} Random Cropping: During training, the ROI of the lips is randomly cropped to an $88 \times 88$  block, whereas center cropping is employed during testing in several lip-reading works~\cite{martinez2020lipreading, petridis2018end}. Flipping: All frames in a video are randomly flipped when a randomly generated probability exceeds 0.5. This data augmentation method is often used in conjunction with random cropping~\cite{martinez2020lipreading, petridis2018end} during training. Mixup: New training examples are created by linearly combining two input video sequences with their corresponding targets. Time Masking \cite{ma2022training}: Randomly masks consecutive frames in each training sequence, replacing the masked frames with either the sequence's average frame or zeros. Word Boundary: After passing through the front-end visual encoder, we often add a word boundary as an additional input to the temporal model. Its length matches the feature dimension length, and vectors corresponding to frames where the target word is present are set to 1, while others are set to 0. After adding the word boundary, the new vector is fed into the temporal sequence model. Label Smoothing: This technique softens the hard values of labels (0 and 1) to address the issue of overconfidence in input labels and enhance model generalization.

\section{METHOD}
\subsection{Framework}

Fig.~\ref{fig1}a illustrates the general framework of our method. Initially, \textbf{InfoSyncNet} employs a 3D convolutional layer to extract temporal and spatial features, resulting in a tensor of shape \(T \times H_1 \times W_1 \times C_1\), where \(C_1\) denotes the number of feature channels. Subsequently, considering the trade-off between accuracy and training complexity, modified ResNet-18 \cite{hara2017learning} is utilized to further refine these features, producing a tensor of shape \(T \times H_2 \times W_2 \times C_2\). The next step involves applying average pooling to condense the spatial knowledge into a shape of \(T \times C_2\). Following the pooling operation (Fig.~\ref{fig1}d), this work introduces an attention mechanism to model the non-uniform information in lip movement sequences and dynamically adapt parameters to different lip inputs. The uniformized information is then input into a Densely Connected Temporal Convolution Network (DC-TCN) for temporal modeling, outputting a shape of \(T \times C_3\) as depicted in Fig.~\ref{fig1}b. This output tensor undergoes another average pooling layer to aggregate temporal information into \(C_4\) channels, where \(C_4\) represents the classes to be predicted. Remarkably, the entire model is designed for end-to-end training. 

\subsection{InfoSyncNet}
Lip reading recognition typically divides the model into a visual front-end encoder and a sequence back-end decoder. The visual front-end is generally used for spatiotemporal feature extraction, while the sequence back-end models the temporal sequences, exploring inter- and intra-sequence relationships. An improved ResNet18 is adopted as the front-end encoder. The original ResNet18 includes one 2D convolutional layer and four residual blocks. The input tensor to our model is \( A \in \mathbb{R}^{T \times H \times W \times 1} \), where $T$ represents the number of time frames in the lip reading sequence, $H$ and $W$ denote the height and width of the frames, and $1$ indicates the channel dimension. To accommodate video data, the first 2D CNN layer of ResNet18 is modified to a 3D CNN layer, resulting in our output tensor \( A^{\prime} \in \mathbb{R}^{T \times H_1 \times W_1 \times C_1} \). The aforementioned process can be described by the following equation:
\begin{equation}A^{\prime} = 3DConv(A). \end{equation}

In addition to the initial modifications, to enhance the convolutional layers' ability to extract and analyze key features across different channels in lip reading sequence images, an SE layer is integrated into each subsequent Residual Block, forming an $SEResidual Block$. Four such operations are denoted as $SEResidual Block$\(_4\). Following this front-end structure, the output dimensions are \(T \times H_2 \times W_2 \times C_2\). Subsequently, a spatial dimension Global Average Pooling ($GAP$) operation is employed to average the features across the entire spatial domain, ensuring the model maintains performance amid positional shifts or changes in image dimensions, while also reducing the number of parameters in the model, thus decreasing computational complexity and the risk of overfitting. After global average pooling, the model output tensor becomes \( B \in \mathbb{R}^{T \times C_2} \). The aforementioned process can be described by the following equation: 
% \begin{equation}B = \mathrm{GPA}(\mathrm{SE-Residual Block\(_4\)}(A^{\prime})).\end{equation}
\begin{equation}B=GAP(SEResidualBlock_4(A^{\prime})).\end{equation}

Subsequently, the backend sequence modeling structure employs a similar DC-TCN~\cite{martinez2020lipreading} architecture as the original backend sequence framework. In the DC-TCN, globally shared convolutional kernels significantly enhance temporal modeling capabilities. However, we observe that the dense connections \cite{huang2017densely} in this model reduce its ability to handle non-uniform information for two main reasons. Firstly, simple models are more effective with highly non-uniform data distributions, as they avoid overfitting to noise and irrelevant patterns. The dense connections in DC-TCN increase the complexity of feature layers—each layer accumulates all features from the input to its current position, potentially reducing the efficiency of the model in integrating these complex features, as reflected in the slower convergence rates of the DC-TCN model~\cite{martinez2020lipreading}. Secondly, dense connections may diminish the network's flexibility in dynamically adapting to varied data features, as evident from DC-TCN's notably poorer performance on the LRW1000 dataset. To address these issues, we introduce an attention module consisting of six Transformer encoder layers between the visual encoder and DC-TCN. It processes high-level features with imbalanced information.

The frontend output $B$ serves as the input to the transformer encoder layer. The computation process of the transformer encoder layer can be represented by the following equations:
\vspace{-1mm}  % 在公式前缩小上方距离
\begin{align}
B^{\prime l} &= B^{l} + \mathrm{MHSA}\left(\mathrm{LN}\left(B^{l}\right)\right), \\
B^{l+1} &= B^{\prime l} + \mathrm{MLP}\left(\mathrm{LN}\left(B^{\prime l}\right)\right).
\end{align}
% \vspace{-1mm}  % 在公式前缩小上方距离
In this context, $l$ = \{1, 2, 3, \ldots, $L$\} and $L$ denotes the number of layers, $B^l$ represents the input to the $l$-th layer, and $\text{LN}(B^l)$ denotes the layer normalization of $B^l$, standardizing the output. $\text{MHSA}$ stands for the Multi-Head Self-Attention mechanism, where the feature dimension is divided into $h$ parts, representing $h$ heads. Self-attention is performed independently on each head, followed by the concatenation of all feature representations along the channel dimension on each head. This division supports parallel processing of information to enhance computational efficiency and model capacity. Moreover, each attention head can focus on different parts of the input sequence, helping the model to capture various complex relationships within and between lip reading sequences across different representational spaces. In each attention mechanism, each partitioned matrix is transformed into a series of queries, keys, and values through the linear transformations. These query, key, and value matrices are packed into matrices $Q$, $K$ and $V$. The similarity between different time frames of each video sequence frame is obtained by multiplying each $Q$ and $K$ matrix. We visualized the similarity matrix obtained by multiplying $Q$ and $K$, which is presented in Subsection \ref{ablationstudy}. After normalization, these attention scores are multiplied with the $V$ matrix, facilitating dynamic feature extraction and contextual integration of the input data. Finally, the outputs from each head are concatenated along the channel dimension to obtain the output of MHSA. Below is the specific calculation formula for the MHSA where $d_k$ is the key dimensionality and $W$ is parameter matric: 
\begin{align}
Att(Q,K,V) &= \mathrm{softmax}\left(\frac{QK^T}{\sqrt{d_k}}\right)V, \\
MHSA(Q,K,V) &= \mathrm{concat}(Att_1,...Att_h)W.
\end{align}

To further enhance feature propagation, a Multi-Layer Perceptron (MLP) is introduced, consisting of two linear transformations connected by a ReLU activation function. The two linear transformations, one for increasing the dimension and one for reducing it, improve information flow and lead to better performance. The specific formula for the MLP operation is as follows:
\begin{align}
Z&=\mathrm{LN}\left(B^{\prime l}\right), \\
\mathrm{MLP}(Z)&=\max(0,ZW_1+b_1)W_2+b_2. 
\end{align}

In addition, the dimensions of both the input and output remain unchanged at $T \times C_2$ for each layer's MHSA and MLP operations. To optimize the training process and improve the model's efficiency and stability, residual skip connections are introduced between layers, enhancing feature reuse across layers. Finally, after $L$ iterations, the final output tensor \( C \in \mathbb{R}^{T \times C_2} \). 

After sequential processing through $L$ layers of the transformer encoder, these optimized features are fed into the DC-TCN network. The output tensor $C$ serves as the current input tensor. First, the input feature information passes through a transition layer ($Tr$), primarily designed for dimensionality transformation. After passing through the transition layer, the dimensions are restored to the desired size, and subsequent transition layers are used for dimensionality reduction. The output tensor $C^{\prime}$ is obtained as follows:
\begin{equation}C^{\prime}=Tr(C)=prelu(BN(Conv1d(C))),\end{equation}
where $BN$ denotes batch normalization. Subsequently, the features will pass through a Dense Block ($DB$), and the following formula describes the detailed process:
\begin{align}
C^{\prime\prime}&=\mathrm{concat}(\mathrm{TC}(\mathrm{SE}(\mathrm{C}^{\prime})),\mathrm{TC}(\mathrm{SE}(\mathrm{C}^{\prime})),\mathrm{C}^{\prime}), \\
C^{\prime\prime\prime}&=\mathrm{concat}(\mathrm{TC}(\mathrm{SE}(\mathrm{C}^{\prime\prime})),\mathrm{TC}(\mathrm{SE}(\mathrm{C}^{\prime\prime})),\mathrm{C}^{\prime\prime}),\\
D&=\mathrm{concat}(\mathrm{TC}(\mathrm{SE}(\mathrm{C}^{\prime\prime\prime})),\mathrm{TC}(\mathrm{SE}(\mathrm{C}^{\prime\prime\prime})),\mathrm{C}^{\prime\prime\prime}),
\end{align}
where $\mathrm{C}^{\prime\prime}\in\mathrm{R}^{\mathrm{T} \times (\mathrm{C_2}+2\mathrm{C_0})}$, $\mathrm{C}^{\prime\prime\prime}\in\mathrm{R}^{\mathrm{T} \times (\mathrm{C_2}+4\mathrm{C_0})}$,
$\mathrm{D}\in\mathrm{R}^{\mathrm{T} \times (\mathrm{C_2}+6\mathrm{C_0})}$. TC denotes a Temporal Convolution layer, and SE refers to the Squeeze-and-Excitation block \cite{hu2018squeeze}, which introduces channel-wise attention to analyze key features across different channels in lip reading sequence images. $C_0$ represents the growth rate, and concat denotes the concatenation operation, which embodies the concept of dense connectivity, used to enhance feature propagation and mitigate the problem of vanishing gradients.

The transition mentioned above and dense block operations will be repeated subsequently, with the specific formula as follows:
\begin{align}
     G=DB(Tr(DB(Tr(DB(Tr(DB(Tr(C)))))))),
\end{align}
where $\mathrm{G}\in\mathrm{R}^{\mathrm{T} \times (\mathrm{C_2}+6\mathrm{C_0})}$, $C_3=C_2 + 6C_0$. After completing the sequence modeling, the current features are processed through a temporal dimension Global Average Pooling ($GAP$), resulting in an output tensor $\mathrm{H}\in\mathrm{R}^{\mathrm{1} \times \mathrm{C_3}}$. Finally, the features are transformed through a linear layer to match the number of classification categories, producing an output tensor $\mathrm{I}\in\mathrm{R}^{\mathrm{1} \times \mathrm{C_4}}$, where $C_4$
represents the final number of classification categories. The specific formulas are as follows:
\begin{align}
    H&=GAP(G),\\
    I&=Linear(H).
\end{align}

\subsection{Loss Function and Training Strategies}
The proposed method employs a cross-entropy loss function for overall optimization, incorporating label smoothing to reduce the model’s overfitting on outlier samples. Given an input sample belonging to category i, where $p_i$ represents the predicted probability and $y$ represents the annotated word label, with $N$ denoting the categories and $q_i$ representing the true label, the cross-entropy loss for multi-class problems can be expressed as:
\begin{equation}\label{eq1}
  L=-\sum_{i=1}^Nq_ilog(p_i)\begin{cases}q_i=0,y\neq i\\q_i=1,y=i,\end{cases}
\end{equation}
After applying label smoothing, the true label $q_i$ is modified as follows:
\begin{equation}\label{eq1}
  q_i=\begin{cases}\epsilon/N&,y\neq i\\1-\frac{N-1}{N}\epsilon&,y=i,\end{cases}
\end{equation}
where $q_i$ no longer an absolute 0 or 1 but is set to a value close to 0 or 1, reducing the model's tendency to overfit to outliers in the training data and $\epsilon \in (0, 1)$. By decreasing the label value on the correct class by the smoothing factor and assigning some small probability values to the incorrect classes, the model learns the boundaries between target classes more smoothly, thereby improving generalization ability.
Additionally, for data augmentation, we utilize Mixup and Time Masking \cite{ma2022training} to increase the diversity of training data, enhancing the model's adaptability in real-life situations. Finally, a word boundary is added between the Attention Module and the DC-TCN, benefiting to more precise lip reading localization.

\section{EXPERIMENT}
\subsection{Setup}
\rz{We evaluate proposed InfoSyncNet against several leading baseline methods, including R18 + BiGRU \cite{feng2020learn}, R18 + DC-TCN / Training strategy \cite{ma2022training} and MTLAM \cite{yeo2023multi}. All experiments are conducted on the most significant lip reading datasets, LRW~\cite{chung2017lip} and LRW1000~\cite{yang2019lrw}.}

\textbf{Dataset.} In this study, we utilized the LRW \cite{chung2017lip} and LRW1000 \cite{yang2019lrw} datasets for our experiments. The LRW dataset comprises English video clips from BBC programs, featuring over 1,000 speakers and a total of 488,766 training samples, along with 25,000 validation and test samples each. The LRW1000 dataset, primarily containing Mandarin content, includes over 2,000 speakers and 1,000 distinct words or phrases. It encompasses 718,018 training, 63,235 validation, and 51,588 test video clips.
% In this study, we employed the LRW \cite{chung2017lip} and LRW1000 \cite{yang2019lrw} datasets for our experiments. The LRW dataset consists of English video clips from BBC programs, featuring over 1,000 speakers, with a total of 488,766 training samples, 25,000 validation samples, and 25,000 test samples. The other dataset, LRW1000, primarily involves Mandarin content and covers over 2,000 speakers with 1,000 different words or phrases. The training, validation, and test sets of this dataset contain 718,018, 63,235, and 51,588 video clips, respectively.

\textbf{Implementation Details.} Our model was trained on a server with two 32GB V100 GPUs over 120 epochs, using a batch size of 64. The optimizer used was AdamW \cite{loshchilov2017decoupled}, with an initial learning rate set at 3e-4 and weight decay at 1e-2. Before training, video samples were randomly shuffled and resized to 96$\times$96, then cropped to 88$\times$88 for model input. Data augmentation included horizontal flipping, grayscale conversion, Mixup, and TimeMasking (TM) \cite{ma2022training}, enhancing generalization.  The $\alpha$ parameter in Mixup was set to 0.2. Based on prior experimental experience, the TM strategy was used only on the LRW dataset. Finally, label smoothing technology was incorporated during the computation of the cross-entropy loss, with the $\epsilon$ parameter set to 0.95, to reduce model overfitting to labels during training. Settings such as word boundary, label smoothing, Mixup, TM, and the Squeeze-and-Excitation block \cite{hu2018squeeze} in the front-end convolution block were all utilized in the methodology of this paper. 

\subsection{Main Results}\label{result}
%表格越界
\begin{table}[ht]
\centering
\caption{Comparison with state-of-the-art methods on the LRW Dataset} 
% \rzc{Give all one abbreviation}
\label{tab:accuracyLRW}
\begin{threeparttable}
\begin{tabular}{
  l
  S[table-format=2.1]
  S[table-format=2.1]
}

\toprule
\textbf{Method} & {\textbf{ Top-1 Accuracy}}  \\
\midrule
P3D R50 + BiLSTM \cite{xu2020discriminative} & 84.8 \\
R18 + BiGRU \cite{feng2020learn} & 88.4 \\
R18 + BiGRU / Face Cutout \cite{Yang2020} & 85.0 \\
R18 + MS-TCN \cite{martinez2020lipreading} & 85.3  \\
R18 + DC-TCN \cite{ma2021lip} & 88.4 \\
R18 + MS-TCN / Born-Again \cite{ma2021towards} & 87.9  \\
R18 + MS-TCN / WPCL + APFF \cite{tian2022lipreading} & 88.3  \\
R18 + MS-TCN / MVM \cite{kim2022distinguishing} & 88.5 \\
EfficientNetV2-L + TransformerTCN \cite{koumparoulis2022accurate}& 89.5 \\
R18 + DC-TCN / Training strategy \cite{ma2022training} & 91.6\tnote{*} \\
MTLAM \cite{yeo2023multi} & 91.7 \\
R18 + MS-TCN / NetVLAD \cite{yang2022improved} & 89.4 \\
3DResNet18 + SA + BiLSTM \cite{axyonov2024audio} & 83.5  \\
ResGNet + C-TCN \cite{jiang2024gslip} & 89.1\\
% \hdashline % Adds a dashed line
InfoSyncNet& \bfseries 92.0 \\
\bottomrule

\end{tabular}
\begin{tablenotes}
\item[*] Ensemble performance of 4 trained models.
\end{tablenotes}
\end{threeparttable}
\vspace{-3mm} % 减少表格内容和下文之间的间距
\end{table}

\begin{table}[ht]
\centering
\caption{Comparison with state-of-the-art methods on the LRW1000 Dataset} % \rzc{ICASSP suggests to supp the result of ref[14]}
% \rzc{Give all one abbreviation}
\label{tab:accuracyLRW1000}
\begin{tabular}{
  l
  S[table-format=2.1]
  S[table-format=2.1]
}
\toprule
\textbf{Method}  & {\textbf{Top-1 Accuracy}} \\
\midrule
R18 + BiGRU \cite{feng2020learn}& 55.7 \\
R18 + BiGRU / Face Cutout \cite{Yang2020} & 45.2 \\
R18 + MS-TCN \cite{martinez2020lipreading} & 41.4 \\
R18 + DC-TCN \cite{ma2021lip} & 43.7 \\
R18 + MS-TCN / Born-Again \cite{ma2021towards} & 46.6 \\
R18 + MS-TCN / MVM \cite{kim2022distinguishing}  & 53.8 \\
MTLAM \cite{yeo2023multi}  & 54.3 \\
ResGNet + C-TCN \cite{jiang2024gslip} & 47.5 \\
% \hdashline % Adds a dashed line
InfoSyncNet & \bfseries 60.7 \\
\bottomrule
\end{tabular}
\vspace{-3mm} % 减少表格内容和下文之间的间距
\end{table}
Table \ref{tab:accuracyLRW} and Table~\ref{tab:accuracyLRW1000} presents the quantitative results on the LRW and LRW1000 datasets and R18 refers to ResNet18. Our method achieves state-of-the-art performance, especially on the LRW1000 dataset. Compared to the current state-of-the-art models, our method improves performance by 0.3\% on the LRW dataset and 5.0\% on the LRW1000 dataset. These results demonstrate our model's enhanced capability in addressing non-uniform visual sequence modeling challenges, attributed to the integration of the attention module. 
\begin{figure}[htb]
    \centering
    \includegraphics[width=1.0\linewidth]{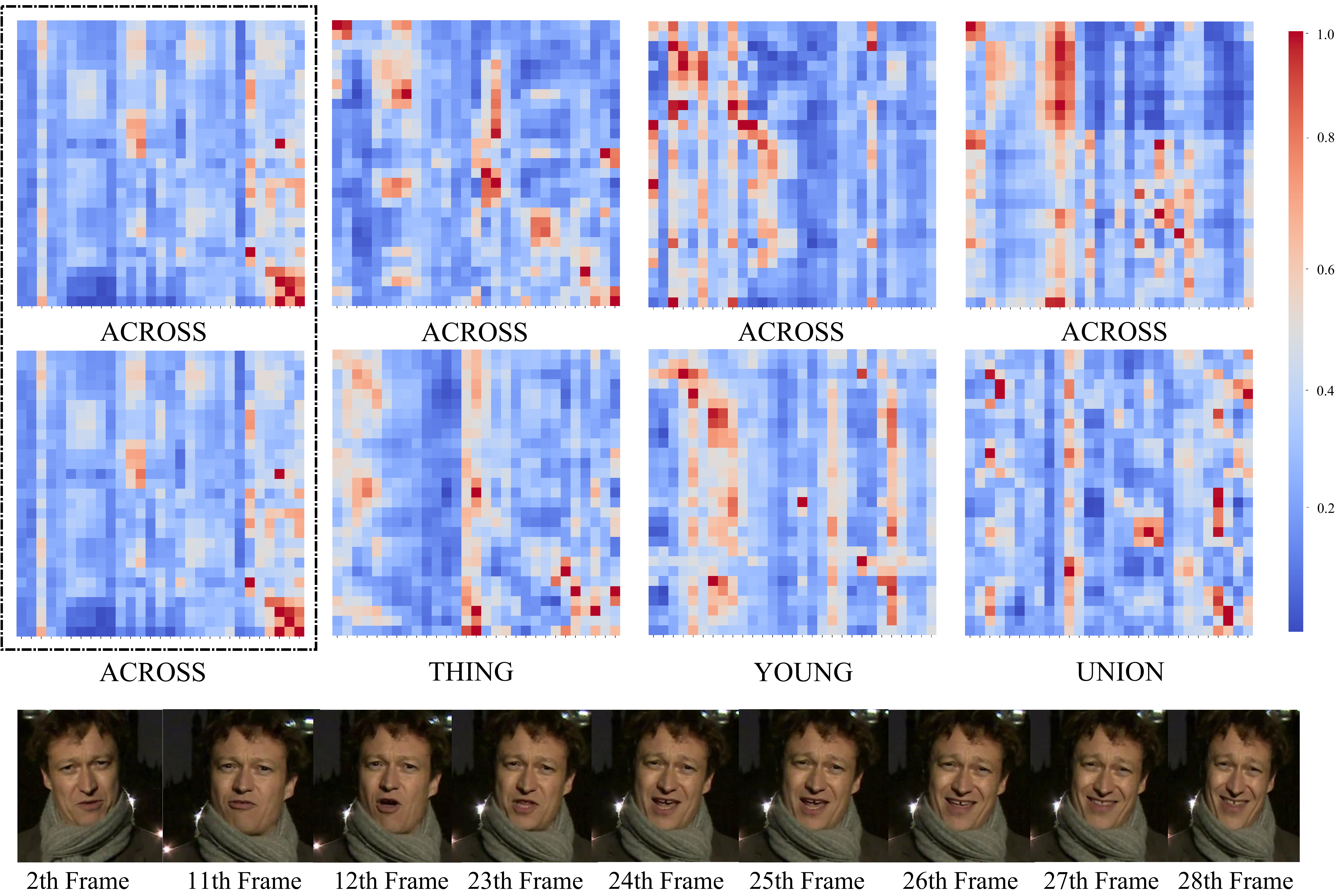}
    \caption{Frame-to-frame correlation heatmap. The horizontal axis represents frames 1-29 of the Key matrix, and the vertical axis corresponds to frames 1-29 of the Query matrix. The first row illustrates correlations from different individuals pronouncing ``ACROSS'', the second from various words, and the third highlights frames identified as significant by the attention mechanism, corresponding to the heatmaps outlined with black dashed lines.}
    \label{fig2}
    \vspace{-5mm} % 减少表格内容和下文之间的间距
\end{figure}
\begin{figure*}[thb]
    \centering
    \includegraphics[width=0.75\linewidth]{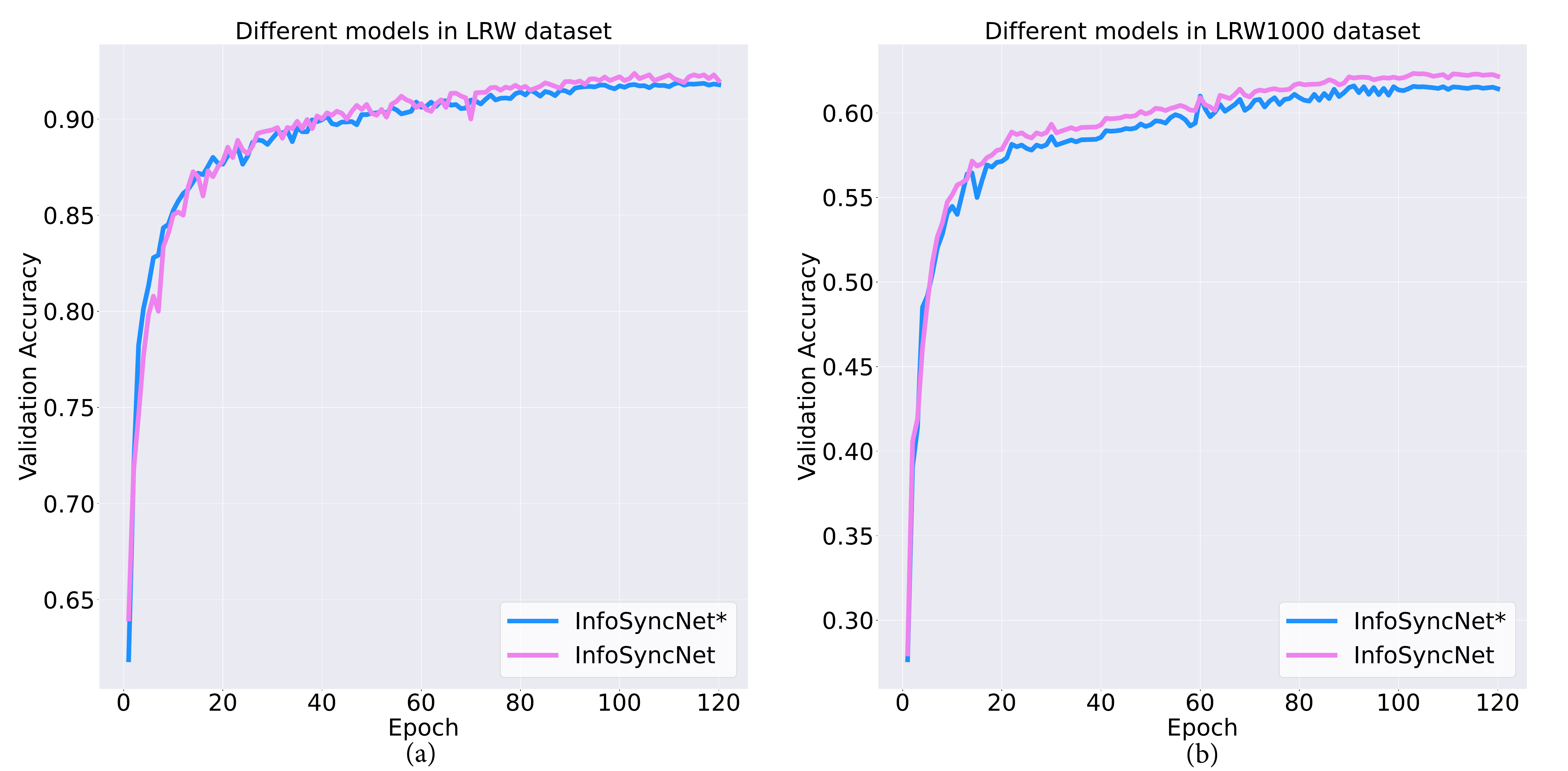}
    \caption{The validation accuracy of the baseline model and the proposed model on two datasets.}
    \label{fig3}
    \vspace{-4mm} % 减少表格内容和下文之间的间距
\end{figure*}

\subsection{Ablation Study}\label{ablationstudy}  To validate the effectiveness of InfoSyncNet, we validate the capabilities of the Transformer encode layer in InfoSyncNet to address cross-sequence diversity and intra-sequence non-uniformity,  which show higher heat values for the frames that receive more attention, as shown in Fig. \ref{fig2}. In the heatmap, the vertical axis represents frames 0-29 of the query matrix, and the horizontal axis corresponds to frames 0-29 of the key matrix. Therefore, columns with higher heat indicate that the corresponding frames are key frames, as they serve as keys related to many queries. 

The first row illustrates how InfoSyncNet manages cross-sequence diversity resulting from different speakers. We visualized the frame-to-frame correlation heatmaps for the word ``ACROSS" across four videos, each from a different speaker. It illustrates the attention module's dynamic adaptation to variations in speaking habits and speeds among different individuals saying the same word. The second row demonstrates InfoSyncNet's handling of cross-sequence diversity introduced by different words. We visualized the frame-to-frame correlation heatmaps for four different words: ``ACROSS", ``THING", ``YOUNG" and ``UNION". The results indicate that InfoSyncNet dynamically adjusts the network’s focus to address cross-sequence diversity and applies varying weights to different frames within the same video to manage intra-sequence non-uniformity. Observing all the heatmaps, it is evident that only a few frames receive primary focus, confirming that lip movements form a non-uniform sequence, which InfoSyncNet addresses by filtering out key information through its attention module. The last row consists of frames that correspond to the primarily focused areas in the highlighted heatmap, capturing significant transitions in lip formations, consistent with human observations. 

\begin{table}[h]
\small
  \caption{Ablation study of the transformer encode layers (Attention Module). InfoSyncNet* denotes InfoSyncNet with the Transformer encoder layers removed}
  \label{tab:freq}
   
  % \vspace{-0.35cm}
  \begin{tabular}{cccc}
    \toprule
Model&Attention Module&LRW(\%)&LRW1000(\%)\\
    \midrule
   InfoSyncNet* & $\times$ & 91.25 & 59.54\\
    InfoSyncNet & $\checkmark$ & 91.98 & 60.72\\
    Improvement  &   &  $\uparrow0.73$ &  $\uparrow1.18$\\
   
  \bottomrule
\end{tabular}
\vspace{-0.3cm}
\end{table}

\begin{table}[h]
\centering
\caption{Ablation studies of diverse training strategies on LRW dataset. ``WB'' stands for Word Boundary, and ``LS'' stands for Label Smoothing}
\label{tab:ablation}
\small % makes the font smaller to fit the column
\begin{tabular}{@{}lcccccc@{}} 
\toprule
Temporal & \multicolumn{2}{c}{Data Augmentation} & WB  & LS  & Top-1  \\
\cmidrule(r){2-3}
Model & TM & Mixup &  &  & Acc.(\%) \\ 
\midrule
InfoSyncNet & $\checkmark$ & $\checkmark$ & $\checkmark$ & $\checkmark$ & 91.98 \\
 & $\times$ & $\checkmark$ & $\checkmark$ & $\checkmark$ &  90.73\\
 & $\checkmark$ & $\times$ & $\checkmark$ & $\checkmark$ & 91.77 \\
 & $\checkmark$ & $\checkmark$ & $\times$ & $\checkmark$ & 89.74 \\
 & $\checkmark$ & $\checkmark$ & $\checkmark$ & $\times$ & 91.55 \\
\bottomrule
\end{tabular}
\vspace{-3mm} % 减少表格内容和下文之间的间距
\end{table}
\begin{table}[h]
\centering
\caption{Ablation studies of diverse training strategies on LRW1000 dataset}
\label{tab:ablationLRW1000}
\small % makes the font smaller to fit the column
\begin{tabular}{@{}lcccccc@{}} 
\toprule
Temporal & \multicolumn{2}{c}{Data Augmentation} & WB  & LS  & Top-1  \\
\cmidrule(r){2-3}
Model & TM & Mixup &  &  & Acc.(\%) \\ 
\midrule
InfoSyncNet & $\checkmark$ & $\checkmark$ & $\checkmark$ & $\checkmark$ & 58.88 \\
 & $\times$ & $\checkmark$ & $\checkmark$ & $\checkmark$ & 60.72 \\
 & $\times$ & $\times$ & $\checkmark$ & $\checkmark$ & 59.96 \\
 &$\times$ & $\checkmark$ & $\times$ & $\checkmark$ & 51.55 \\
 & $\times$ & $\checkmark$ & $\checkmark$ & $\times$ & 58.37 \\
\bottomrule
\end{tabular}
\vspace{-3mm} % 减少表格内容和下文之间的间距
\end{table}
In addition to the qualitative experiments, we conducted quantitative analyses of the attention module, as shown in the ablation study below in Table~\ref{tab:freq} and in Fig. \ref{fig3}. InfoSyncNet* denotes InfoSyncNet with
the Transformer encoder layers removed. The numbers represent top-1 accuracy and compared to the model without Transformer encoder layers, our method improves performance by 0.73\% on the LRW dataset and 1.18\% on the LRW1000 dataset. Additionally, to validate the effectiveness of our training strategies, we conduct extensive experiments to optimize a combination that balances effectiveness with efficiency, as detailed in Table~\ref{tab:ablation} and Table~\ref{tab:ablationLRW1000}. From Tables 4 and 5, Word Boundary (WB) emerges as the most effective training strategy, followed by Label Smoothing (LS). Additionally, an interesting observation is that Time Masking benefits the LRW dataset but adversely affects the LRW1000 dataset. Time Masking, as a data augmentation technique, enhances the model's ability to process partial information in single-view lip-reading datasets like LRW, resulting in positive effects. However, in multi-view datasets such as LRW1000, it may impair the model's ability to understand cross-view temporal information, leading to decreased performance. Therefore, tailoring and optimizing the Time Masking strategy based on the specific characteristics of the dataset is crucial for improving model performance. These strategies address challenges posed by diverse lighting conditions and speaker orientations while maintaining computational efficiency.

\section{CONCLUSION}
We present InfoSyncNet, a novel non-uniform sequence modeling network that effectively handles both cross-sequence diversity and intra-sequence non-uniformity. The core design of InfoSyncNet is a non-uniform quantization module strategically positioned in the feature layer, which provides the temporal modeling component with more uniformly distributed input. Ablation studies, utilizing visualization of the attention module's functionality, confirm our core contribution and comprehensive comparative experiments on relative benchmark datasets demonstrate InfoSyncNet’s effectiveness in Visual Speech Recognition. Lastly, we also conduct ablation experiments to compare the impact of various training strategies on lip-reading tasks.

\bibliographystyle{IEEEtran}
\bibliography{IEEEabrv,mylib}
\end{document}